# Individual Survival Curves with Conditional Normalizing Flows

Guillaume Ausset*[†], Tom Ciffreo[†], Francois Portier*, Stephan Clémençon*, Timothée Papin[†]
*Télécom Paris, Saclay, France
name.surname@telecom-paris.fr
[†] BNP Paribas, Paris, France
name.surname@bnpparibas.com

*Abstract*—Survival analysis, or time-to-event modelling, is a classical statistical problem that has garnered a lot of interest for its practical use in epidemiology, demographics or actuarial sciences. Recent advances on the subject from the point of view of machine learning have been concerned with precise *per-individual* predictions instead of population studies, driven by the rise of individualized medicine. We introduce here a conditional normalizing flow based estimate of the time-to-event density as a way to model highly flexible and individualized conditional survival distributions. We use a novel hierarchical formulation of normalizing flows to enable efficient fitting of flexible conditional distributions without overfitting and show how the normalizing flow formulation can be efficiently adapted to the censored setting. We experimentally validate the proposed approach on a synthetic dataset as well as four open medical datasets and an example of a common financial problem.

## I. INTRODUCTION

It is the purpose of this paper to show how the recent advances on normalizing flows can be transposed to the survival analysis setting. We show how normalizing flows i.e. bijective maps from a known and simple distribution to a target distribution of interest, provide a natural framework for the modelisation of complex conditional survival distributions that need to be easily sampled from, for example if complex processes need to be simulated. We propose the first, to the best of our knowledge, normalizing flows approach to survival analysis. We show how to model a time-to-event, i.e. a positive random variable $T$ which may be only partially observed, conditioned upon some covariates $X \in \mathbb{X}$. In the standard survival analysis setting, the quantity $T$ of interest is considered to be right-censored, meaning we suppose the existence of some nuisance random variable such that only imperfect data is observed. This setting is widely assumed in medicine, predictive maintenance or finance, where individuals are followed through their lifetime in order to record a certain event of interest (death, malfunction or credit default for example) that can eventually never be observed because of (potentially random) external factors such as the end of the study or any event that invalidates any later observations. As will be detailed in section II, multiple quantities can be equivalently modelled in order to uniquely define the distribution of the survival times; approaches based on modelling the hazard rate under a proportionality hypothesis yield the Cox estimator [1], which has been extended to accommodate increasingly flexible families of functions such as in [2] where a deep neural network parametrizes the proportional odds. More generally the proportional hazards hypothesis can be weakened by allowing dependencies on time, such as in [3] or by the use of a finite mixture such that the proportional assumption holds conditionally, such as in [4]. It is possible to entirely discard the proportional hazard hypothesis provided one is able to solve a differential equation, this is the approach taken in [5] where the authors solve the problem using Neural ODEs (see [6]). Similarly, [7] uses Gaussian Processes to model the instantaneous hazard rate. Direct modelling of the density of event times is also possible, either by parametrization of a known parametric model, which is the approach taken by [8] where a neural network parametrizes a deep exponential family and in [9] where a neural network is used to parametrize the parameters of a Weibull. These approaches, while able to encode complex relations in the data by means of neural networks, are still constrained by the choice of parametric family. Attempts to model the density without such restrictive assumptions have been developed, [10], [11] directly model the density without making any assumption on the distribution but do so by discretizing the space, an undesirable limitation if the estimator needs to be used as a plugin. In this paper we instead directly model the density as the pushforward measure of a simpler, and known, measure by modelling it as a normalizing flow. Compared to the similar discrete normalizing flows approach of [12], we gain the ability to directly train using the log-likelihood without having to rely on an approximate variational approach or giving up the ability to efficiently draw samples. Our factored conditional representation also greatly reduces the computational cost and gives us the ability to go beyond the single layer used by the authors. Similarly to the approaches of [5], [10], [12], we do not have to make any restrictive assumptions on the form of the distribution, but unlike all the previous approaches we are able to efficiently sample from the distribution of event times. Finally, compared to the DeepSurv or ODE approaches, we gain the ability to exactly compute some useful statistics such as the quantiles by delegating the computation to the simpler and known pullback distribution. Normalizing flows have been successfully applied to unconditional density estimation and generative modelling in various applications since their inception but have recently been adapted to the conditional setting; [13] uses conditional flows in order to solve the motion planning problem by modelling the policy, while [14] applies conditional normalizing flows to the image generation problem. While these approaches integrate conditioning in the normalizing flow framework, they do not show how to efficiently deal with the potentially high-dimensional covariates, choosing to simply pass it as auxiliary input to the flow. We instead show here how a simple factorization scheme can be used to ease the computational burden and how to use the structure of the problem to add different levels of conditioning at different steps.

In section II we introduce the probabilistic background necessary to understand the survival analysis setting while in section III we show how normalizing flows can straightforwardly be applied to the estimation of censored distribution by transporting the computation of the survival back to the latent space. In section IV we give a simple scheme to introduce conditioning in the flow while keeping the computational cost reasonable. Finally, in section V we show that the approach proposed matches or outperforms current state-of-the-art techniques both on synthetic and real medical datasets, and we give in subsection V-C an example of the usefulness of efficient sampling from the learned distribution through a toy example inspired by a common financial situation.

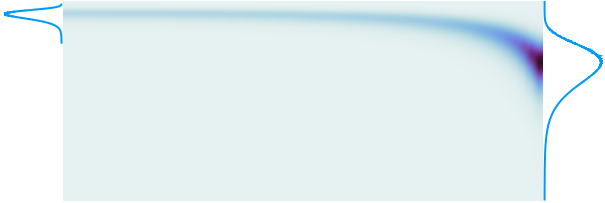

Figure 1. Mapping a normal distribution to a survival distribution.

## II. Background

We consider a positive random variable $T \in \mathbb{R}^+$ which may be only partially observed and following some distribution $\mathcal{T}$, conditioned upon some covariates $X \in \mathbb{X}$. We place ourselves in the standard right-censored setting, meaning we suppose the existence of some nuisance random variable $C \in \mathbb{R}^+$ such that the observed data consists only of the triplet $(Y, \delta, X)$, where $Y = \min(T, C)$ is the observed time and $\delta = \mathbb{1}_{T \leq C}$ is the censoring indicator where $\mathbb{1}_x(y)$ is the indicator function, defined by the relation

$$\mathbb{1}_x(y) = \begin{cases} 1 & \text{if } x = y \\ 0 & \text{otherwise.} \end{cases}$$

Other forms of censoring or truncation exist but are not considered here. Do note, however, that the results can easily be adapted.

The study of the survival conditional distribution of $T \mid X$, the times to events (ignoring the issues of identifiability caused by the loss of information in the tail when no hypothesis is imposed on $C$) can be addressed in different manners depending on the primary quantity that is being modelled. The particular structure of the right censored problem enables the possibility to model interchangeably either the density $f_T(\cdot \mid X = x)$ of $T \mid X = x$, its survival function $S_T(\cdot \mid X = x)$ defined by

$$S_T(t \mid X = x) = \int_t^\infty f_T(u \mid X = x) du,$$

the instantaneous hazard rate $h_T(\cdot \mid X = x)$ or the integrated hazard rate $H_T(\cdot \mid X = x)$, defined by:

$$h_T(t \mid X = x) = \lim_{dt \to 0} \frac{\mathbb{P}(t \leq T \leq t + dt \mid t \leq T, X = x)}{dt}$$

$$H_T(t \mid X = x) = \int_0^t h_T(u \mid X = x) du.$$

Any of the previous four quantities fully characterizes the conditional law of $T$ given $X$ and can be used to recover the other three through the identities

$$\frac{dS_T(t \mid X = x)}{dt} = -h_T(t \mid X = x) S_T(t \mid X = x) \quad (1)$$

$$S_T(t \mid X = x) = \exp(-H_T(t \mid X = x)).$$

The instantaneous hazard rate is usually considered the most natural quantity insofar as it can be directly interpreted: it represents the instantaneous probability of an event happening now as opposed to the density $f_T$ which represents the instantaneous probability of an event happening seen from the origin $T = 0$. Choosing to estimate $h_T$ is therefore a sensible choice and using the previous estimator as a plugin estimator in (1) yields the time-honoured nonparametric Nelson–Aalen and Kaplan-Meier estimators. Those nonparametric estimators (see [15] for the conditional version) are, however, highly susceptible to the pitfalls of the curse of dimensionality and therefore prone to overfitting.

Similarly to regression models such as the AFT model, we can assume that $T \sim \mathcal{T}$ can be obtained as a transformation by some mapping $g$ of some latent variable $Z \in \mathcal{Z}$ with $Z \sim \mathcal{Z}$ such that

$$T = g(Z, X). \quad (2)$$

This view is not restrictive as we can indeed always write

$$T = F^{-1}_{T\mid X=x} \circ F_Z(Z)$$

when all the variables are continuous as we have $F_Z(Z) \sim \mathcal{U}[0, 1]$ and $F^{-1}_{T\mid X=x}(U) \sim \mathcal{T}$ when $U \sim \mathcal{U}[0, 1]$. As $F_{T\mid X=x}$ is unknown, it is natural to instead select the best candidate $g_{\theta^\star}$ from some family $(g_\theta)_\theta$, parameterized by $\theta \in \Theta$ (usually $\Theta = \mathbb{R}^D$), that minimizes some notion of distance to the true distribution $\mathcal{T}$ i.e. such that

$$g_{\theta^\star} = \arg\min_{\theta \in \Theta} \mathcal{D}(\mu_\mathcal{T}, g_\theta(\mu_Z, X)),$$

where $g_\theta(\mu_Z, X)$ denotes the push-forward measure of $\mu_Z$ with mapping $g_\theta$ and conditioning $X$. While several distances

$\mathcal{D}$ have been proposed in the censored setting, such as CRPS [16] for sharp estimates, we directly use here the negative log-likelihood. It is possible to write in the right censored setting the likelihood of the observed data in terms of relevant quantities only

$$p((Y, \delta, X) \mid \theta) \propto f_{T|X,\theta}(Y)^\delta S_{T|X,\theta}(Y)^{1-\delta}.$$

where $f_{T|X,\theta}(Y)$ is the density of the pushforward measure $g_\theta(\mu_Z, X)$ and $S_{T|X,\theta}(Y)$ its corresponding survival. The previous equation yields the empirical risk minimization problem

$$\arg\min_\theta \frac{1}{n} \prod_{i=1}^n f_{T|X_i}(Y_i)^{\delta_i} S_{T|X_i}(Y_i)^{1-\delta_i}. \quad (3)$$

In the following sections, we will see how to construct the mapping (2) such that the quantities involved in 3 are both computable and automatically differentiable i.e. amenable to differentiation exactly algorithmically (we refer to [17] for an overview of the field of *automatic differentiation*).

III. UNCONDITIONAL SURVIVAL NORMALIZING FLOWS

In this section we momentarily only concern ourselves with the *unconditional* estimation of the survival and omit the conditioning on the covariates $X$. This is only for the sake of readability and one needs only to introduce the missing conditioning in all the following relations to retrieve the conditional case.

A. Computing $f_T$

Assuming the existence of the mapping $T = g_\theta(Z)$ and under the hypothesis that $g_\theta$ is a $C^1$-diffeomorphism it is possible to derive the density of $T$ from the density of $Z$ from the change of variable theorem

$$\log f_T(t) = \log f_Z(z) - \log\left|\det \frac{\partial g_\theta}{\partial z}\right|. \quad (4)$$

Equation (4) imposes not only the explicit constraint that $g_\theta$ must be invertible but also that the determinant of the jacobian is easy to compute. Such constraints are in practice fairly difficult to meet and force $g_\theta$ to be fairly simple with a jacobian whose determinant is efficiently computable. It is, however, possible to retrieve the lost representational power by simply composing multiple simple transformations such that

$$T = g_{\theta,K} \circ \cdots \circ g_{\theta,0}(Z) \quad (5)$$

$$\log f_T(t) = \log f_Z(z) - \sum_{i=0}^K \log\left|\det \frac{\partial g_{\theta,i}}{\partial z_i}\right|$$

Since the original paper of [18] most of the research on the subject has gravitated around constructing families of functions $g_\theta$ that are easily computable and whose jacobian has a tractable determinant i.e. diagonal, triangular, have a simple block structure or more generally to encode an adjacency matrix [19] while maintaining a high degree of flexibility.

It is, however, possible to entirely sidestep the previous problems by defining (5) continuously, as proposed by [6].

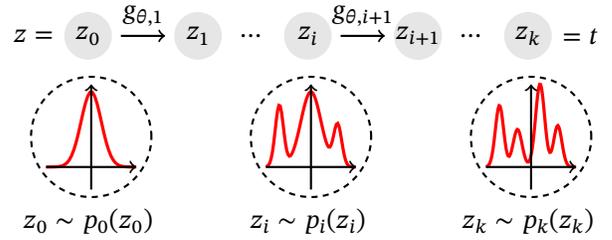

Figure 2. Mapping a simple distribution to a target distribution by successive compositions.

By interpreting the change of variable associated to a single composition step as an Euler integration step from $i$ to $i+1$, we can instead adopt an infinitesimal point of view by parametrizing the derivative of the change of variable, it is possible to prove [6] that the change of variable theorem become:

$$z_{i+1} = g_{\theta,i}(z_i) \qquad \frac{\log f(z_{i+1}) - \log f(z_i)}{i+1-i} = -\log\left|\frac{\partial g_{\theta,i}}{\partial z}\right|$$

$$\frac{\partial \mathbf{z}_\theta}{\partial t} = g_\theta(\mathbf{z}_\theta(t), t)^1 \qquad \frac{\partial \log f(\mathbf{z}_\theta(t))}{\partial t} = -\operatorname{tr}\frac{\partial g_\theta}{\partial \mathbf{z}}$$

We distinguish here, $\mathbf{z}(t)$ the path of the flow, from $z$ the initial latent variable. $\mathbf{z}(t)$ is only a mathematical device used to define the transformation of interest and is not in itself the object of interest.

The compositional process of (5) is therefore replaced by the initial value problem

$$\frac{\partial}{\partial t}\begin{bmatrix} \mathbf{z}_\theta(t) \\ \log f(y) - \log f_{\mathbf{z}_\theta(t)}(\mathbf{z}_\theta(t)) \end{bmatrix} = \begin{bmatrix} g_\theta(\mathbf{z}_\theta(t), t) \\ -\operatorname{tr}\frac{\partial g_\theta}{\partial \mathbf{z}}\bigg|_{\mathbf{z}_\theta(t), t} \end{bmatrix}$$

$$\begin{bmatrix} \mathbf{z}_\theta(1) \\ \log f(y) - \log f_{\mathbf{z}_\theta(1)}(\mathbf{z}(1)) \end{bmatrix} = \begin{bmatrix} y \\ 0 \end{bmatrix}. \quad (6)$$

Note that the problem as written here defines the flow in the direction $\mathcal{T} \to \mathcal{Z}$ i.e the mapping from $T$ to $Z$. The (inverse) mapping $\mathcal{Z} \to \mathcal{T}$ is similarly defined by changing the starting point (and matching initial conditions) of the problem. We denote by $\mathbf{z}_\theta(t, z_1)$ (resp. $\mathbf{z}_\theta(t, z_0)$) the path solution of the particular instance of the IVP (6) with initial condition $\mathbf{z}_\theta(1, z_1) = z_1$ (resp. $\mathbf{z}_\theta(0, z_0) = z_0$) and parametrization $\theta$ of the dynamics. By parameterizing a normalizing flow infinitesimally, we are able to overcome the two previous limitations: not only the expensive computation ($O(n^3)$) of the determinant is entirely eliminated and replaced by a trace operation ($O(n)$) but the restriction on invertibility is not explicitly required anymore: as long as $g_\theta$ and $\frac{\partial}{\partial \mathbf{z}} g_\theta$ are Lipschitz continuous then (6) admits a solution (see [20] or [21]). In practice these hypotheses are met for most of the common layers and activation functions used in deep learning. This solution is also invertible by construction and only requires solving the equation backward in time.

---

[1]One can interpret $g_{\theta,i}(z_i)$ as $g_\theta(z_i, i)$.

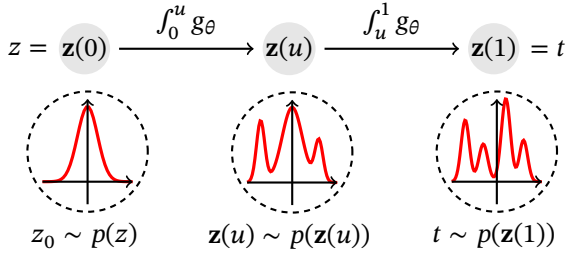

Figure 3. Mapping a simple distribution to a target distribution by continuously applying an infinitesimal flow.

We note here that while $g_\theta$ in the continuous definition of the normalizing flows plays a similar role to $g_\theta$ in the discrete version, it is in fact not the same object and does not represent the flow. Instead, for $\mathbf{z}_\theta(\cdot, z_0)$ solution of (6) with initial value $\mathbf{z}_\theta(0) = z_0$ we denote by $G_\theta$ the resulting normalizing flow such that

$$G_\theta(z) = \mathbf{z}_\theta(1, z_0),$$
$$G_\theta^{-1}(z) = \mathbf{z}_\theta(0, z_1).$$

As it is not possible to ensure that the learned flow maps from $\mathbb{Z} \to \mathbb{R}^+$, where $\mathbb{Z}$ is the support of pullback distribution $\mathcal{Z}$ (considered an hyperparameter to be chosen by the user), we instead reparametrize the normalizing flow using a two-step process. First a learned flow maps from $\mathcal{Z}$ to $\mathbb{R}$ by learning

$$\log T = G_\theta(Z).$$

then a deterministic change of variable maps $\mathbb{R}$ to $\mathbb{R}^+$ through

$$T = \exp(\log T) = \exp\left(G_\theta(Z)\right).$$

The resulting process is still a normalizing flow and could be reframed as a continuous normalizing flow by deriving the corresponding dynamics. This reparameterization ensures that $G_\theta(Z) \in \mathbb{R}^+$ is a proper time-to-event as well as prevents any issues arising from the possibility of mapping an event outside the support of the latent distribution. This reparameterization is common in survival *regression* and our model can be seen as a generalization of the accelerated failure time model of [22]. For simplicity, however, we perform the last change of variable independently and we omit this last step from the notation in the following sections for the sake of readability.

### B. Computing $S_Y$

While the system given previously describes how to obtain $Y_i = G_\theta(Z_i)$ as well $f_T(Y_i) = \log p(G_\theta(Z))$, we also need to be able to compute $S_T(Y_i)$ in order to entirely define the loss (3).

We exploit the relation between $T$ and $Z$ and note that,

$$\begin{aligned}
S_Y(T_i) &= \int_{\mathbb{R}^+} \mathbb{1}_{T \geq T_i} \mathrm{d}(\mu_\mathcal{T}) = \int_{\mathbb{R}^+} \mathbb{1}_{G_\theta(Z) \geq T_i} \mathrm{d}\left(G_\theta\left(\mu_\mathcal{T}\right)\right) \\
&= \int_{\mathbb{Z}} \mathbb{1}_{Z \geq G_\theta^{-1}(T_i)} \mathrm{d}\left(\mu_\mathcal{T}\right) \\
&= S_Z(G_\theta^{-1}(T_i)).
\end{aligned} \quad (7)$$

where the penultimate equality is not trivial but can be obtained by taking the derivative of $G_\theta(z) = z(1, z, \theta)$ with respect to $z$ which yields, provided $g_\theta$ is sufficiently smooth, an adjoint initial value problem whose dynamical system is loosely decoupled, with adjoint state

$$\frac{\mathrm{d}}{\mathrm{d}z} G_\theta(z) = \exp\left(\int_0^1 -\frac{\partial}{\partial z} g_\theta \left(\mathbf{z}_\theta(t, z), t\right) \mathrm{d}t\right) > 0$$

We do not solve the adjoint system but only the positivity information to infer the sense. From (7), we see that computing the survival function on the image space can be reframed as computing the survival of the preimage. By construction, computing $G_\theta^{-1}(T_i)$ is of the same complexity as computing $G_\theta(Z_i)$ and only requires solving the system backward in time.

Now that we have shown the quantities to be *computable*, we show how to efficiently obtain the gradient of these quantities.

### C. Parameter Estimation

We recall that we wish to find $\theta$, solution of the empirical risk minimization problem of Equation (3). As the use of computationally expensive neural networks constrains our choice of optimization methods to first order gradient methods, we only need to be able to differentiate $G_\theta$, solution of the IVP (6).

As all the quantities involved are differentiable, we only need to be able to compute the sensitivity $\partial_\theta G_\theta(Z)$ of the ODE solution itself with respect to its parameters, usually referred in the ODE literature to local sensitivity analysis (as opposed to global sensitivity analysis which concerns itself with the study of the range of solutions given the whole feasible domain of inputs and parameters).

By rewriting $G_\theta(Z)$ as

$$\begin{aligned}
G_\theta(Z) &= \mathbf{z}_\theta(1, \theta) \\
&= \int_{t_0}^T \mathbf{z}_\theta(t, \theta)^\intercal \begin{bmatrix} 0 \\ 1 \end{bmatrix} \delta_1(t) \mathrm{d}t,
\end{aligned}$$

the loss (3), given a solution $\mathbf{u}$ of the IVP (6) (with $\mathbf{u} = [\mathbf{z}_\theta, \Delta \log f(\mathbf{z}_\theta)]$), can be written as

$$L(\mathbf{u}, \theta) = \int_{t_0}^T l(\mathbf{u}(t, \theta), \theta) \mathrm{d}t,$$

we can form the adjoint state

$$\frac{\partial \lambda}{\partial t} = \frac{\partial l}{\partial u}(\mathbf{u}(t, \theta), \theta) - \lambda(t) \frac{\partial f}{\partial u}(t, \mathbf{u}(t, \theta), \theta), \quad \lambda(T) = 0.$$

such that

$$\begin{aligned}
\frac{\partial L}{\partial \theta} = \int_{t_0}^T &\lambda(t) \frac{\partial f}{\partial \theta}(\mathbf{u}(t, \theta), \theta) \\
&+ \frac{\partial l}{\partial \theta}(\mathbf{u}(t, \theta), \theta) \mathrm{d}t + \lambda(t_0) \frac{\partial \mathbf{u}}{\partial \theta}(t_0, \theta).
\end{aligned}$$

This adjoint method (see [23]) doesn't require any additional method (other than a $O(1)$ increase in computations) in order to obtain the derivative and the solution itself, we only need to solve the original IVP with the addition of the new adjoint state, resulting in the same asymptotic computational complexity.

Using the adjoint method, we are therefore able to differentiate $G_\theta = \mathbf{z}_\theta(1, \cdot)$ as well as $G_\theta^{-1} = \mathbf{z}_\theta(0, \cdot)$ with respect to $\theta$. We refer to [24] for a more complete overview of the various possibilities for automatic differentiation of a solution of a differential equation as well as [25] for the perspectives offered.

## IV. CONDITIONAL SURVIVAL NORMALIZING FLOWS

In the previous sections we omitted the conditioning on the covariates $X \in \mathbb{X}$ in order to simplify the notations. We now reintroduce this conditioning, as its presence does not modify any of the previous results, and show how to construct efficiently a mapping $G_\theta \equiv \mathcal{Z} \to \mathcal{T} \mid X$ such that

$$T = G_\theta(Z, X).$$

As the trace operator is linear, it is possible to efficiently extend the expressivity of a single normalizing flow at a minor computational cost by representing it as a linear combination of $K$ basis functions i.e.

$$g_\theta(\mathbf{z}_\theta(t,x), t, x) = \sum_{i=1}^{K} g_{\theta,i}(\mathbf{z}_\theta(t,x), t, x)$$

As in [6], we choose to parametrize each basis function $g_{\theta,i}$ as a mixture of unconditional and time-invariant dynamics. We chose to decouple the gating in $x$ and $t$ in order to prevent overfitting and be able to apply different regularizations and computational budget. Decoupling the blocks in $x$, $t$ and $z$ also gives us the ability to exploit the structure of the problem to implement efficient batching.

$$g_\theta(\mathbf{z}_\theta(t,x), t, x) = \sum_i \pi_{\theta,i}(x)\sigma_{\theta,i}(t)g_{\theta,i}(\mathbf{z}_\theta(t,x))$$

The full dynamics of our continuous normalizing flow are therefore,

$$\frac{\partial}{\partial t}\mathbf{z}_\theta(t,x) = \sum_i \pi_{\theta,i}(x)\sigma_{\theta,i}(t)g_{\theta,i}(\mathbf{z}_\theta(t,x))$$

$$\frac{\partial}{\partial t}\log f(\mathbf{z}_\theta(t) \mid x) = -\sum_i \pi_{\theta,i}(x)\sigma_{\theta,i}(t)\,\mathrm{tr}\left.\frac{\partial g_{\theta,i}}{\partial z}\right|_{\mathbf{z}_\theta(t,x)}.$$

### A. Hierarchical Conditioning

Conditional density estimators learned by maximizing the likelihood are well known to be prone to overfitting. We control the amount of overfitting by introducing an auxiliary latent representation shared by all the conditional distributions $T \mid X$. We therefore impose the shared hierarchical representation $w = H_\theta(z)$ such that $t = G_\theta(w, x) = G_\theta(H_\theta(z), x)$ with $H_\theta(Z) \perp\!\!\!\perp X$. The corresponding flow dynamics can be rewritten as

$$\frac{\partial \mathbf{z}_\theta}{\partial t}(t,x) = \left[\sum_{i=1}^K \left(\mathbb{1}_{t \le t_x} + \pi_{\theta,i}(x)\mathbb{1}_{t > t_x}\right)\sigma_{\theta,i}(t)g_{\theta,i}(\mathbf{z}_\theta(t,x))\right]$$

where $t_x$ controls implicitly the allowed deviation from the unconditional distribution. If we denote by $f_{H_\theta(z)}$ the unconditional distribution induced by $H_\theta(z)$ and $f_{G_\theta(H_\theta(z),x)}$ the distribution induced by $f_{G_\theta(H_\theta(z),x)}$, we can regularize the intermediate shared latent representation toward the true *unconditional survival distribution* by augmenting the loss (3) with an intermediary loss $\mathcal{L}_u$ such that

$$\begin{aligned}\mathcal{L}_{\text{total}}(\theta) &= \mathcal{L}_u(\theta) + \mathcal{L}_c(\theta) \\ &= \sum_i \delta_i \log(f_{H_\theta(Z_i)}(T_i)) \\ &\quad + (1-\delta_i)\log(S_{H_\theta(Z_i)}(T_i)) \\ &\quad + \lambda \sum_i \delta_i \log(G_\theta(H_\theta(Z_i), X_i)(T_i)) \\ &\quad + (1-\delta_i)\log(S_{G_\theta(H_\theta(Z_i),X_i)}(T_i)),\end{aligned}$$

where $S_{H_\theta(x)}$ and $S_{G_\theta(H_\theta(z),x)}$ are the respective survival functions of the distributions defined by the densities $f_{H_\theta(z)}$ and $f_{G_\theta(H_\theta(z),x)}$.

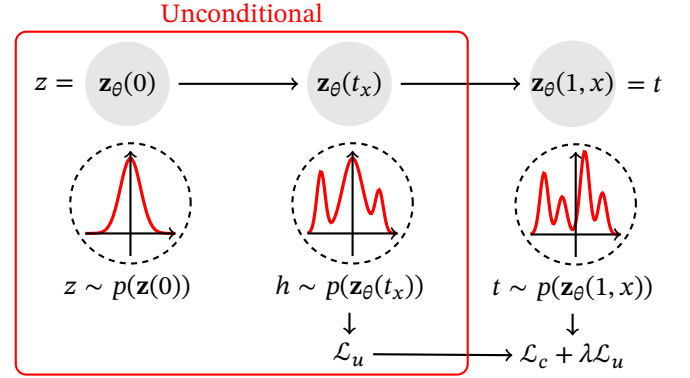

### B. Discrete & Continuous Hierarchical Conditioning

For datasets with particularly complex dependences on the covariates, it is possible to add other layers of hierarchies. The simplest scheme consists of using *discrete* hierarchical transformations: let $H$ be the number of hierarchies then we can learn $K \times H$ mixtures such that

$$\begin{aligned}&g_\theta(\mathbf{z}_\theta(t,x), t, x) \\ &= \sum_{h=1}^{K} \mathbb{1}_{t_{h-1} < t \le t_h} \sum_{i=1}^{K} \pi_{\theta,i,h}(x)\sigma_{\theta,i,h}(t)g_{\theta,i,h}(\mathbf{z}_\theta(t,x)).\end{aligned}$$

We call this scheme *discrete hierarchical conditioning*. In the same manner as done in the previous section, it is possible if desired to introduce intermediary losses in order to prevent overfitting as well as help with the training procedure. As our model is continuous, we do not have to constrain ourselves to hard gating at time steps $(t_h)$, effectively reducing our model to a standard normalizing flow using continuous normalizing

flow layers. Instead we can *continuously* interpolate between representations:

$$g_\theta(\mathbf{z}_\theta(t,x),t,x) = \sum_{h=1}^{H} \exp\left(c_h |t-t_h|^2\right) \sum_{i=1}^{K} \pi_{\theta,i,h}(x) \sigma_{\theta,i,h}(t) g_{\theta,i,h}\left(\mathbf{z}_\theta(t,x)\right).$$

We call this scheme *continuous hierarchical conditioning*. By making $c_h$ and $t_h$ a learnable parameter, we make it possible for the model to learn if hierarchies are needed.

The hierarchical approach as introduced previously may at first appears strictly identical to the non-hierarchical approach as it is possible to incorporate $\mathbb{1}_{t_{h-1} < t \le t_h}$ directly inside $\sigma$. There is, however, one significant advantage: by knowing the relative order (in the sense of $t$) in which the mixtures are applied we can design $g_{\theta,i,h}$ to be of increasing complexity in order to impose a shared representation between the different individuals.

## V. Experiments

All the material, including code and data, necessary for the reproduction of the results presented here is available at `git.sr.ht/~aussetg/nfsurvival`. The experiments have been implemented in the Julia language [26] where we make heavy use of the `DifferentialEquations.jl` [27] and `Zygote.jl` [25] libraries in order to implement automatic differentiation of the initial value problems involved. The normalizing flow approach is directly compared to existing methods for survival analysis on synthetic data designed to model violations of the proportional hazards hypothesis as well as multimodality. We compare our approach to the existing literature on standard open medical datasets and motivate the need for generative models that are easy to sample from by applying our method to a commonly encountered setting in the financial community.

### A. Synthetic Data

In order to test the ability to both capture complex interactions between covariates as well as model a potentially multimodal distribution which violates the proportional hazard assumption, we generate synthetic data according to the following model:

$$X \sim \mathcal{U}_d$$
$$T \sim pW(\beta_1^\mathsf{T} X, \beta_2^\mathsf{T} X) + 0.7 W(2 * \beta_3^\mathsf{T} X, \beta_4^\mathsf{T} X)$$
$$C \sim W(\beta_5^\mathsf{T} X, \beta_6^\mathsf{T} X)$$
$$Y = \min(T, C)$$
$$\delta = \mathbb{1}_{T \le C},$$

with some resulting distributions represented in subsection V-A. While often implicitly discarded by the model chosen, the possibility for the event distribution to be multimodal is far from exotic: many diseases, such as acute radiation poisoning, include a latent period of relative well-being of the patients; death occurring before or after the latent period but not during. Similarly, in a financial setting it is expected to observe modes around important fiscal events as those are the periods one expects a company to default. The different models are trained on a test set of 3000 observations (chosen to match the characteristics of the real datasets) with an average of 80% of censoring. We then estimate the Harrel's concordance index [28], [29]

$$\mathbb{P}_Y\left(s(X_i) \ge s(X_j) \mid T_i \le T_j\right)$$

where $s(X_i)$ is a scoring function defined later and $(X_i, T_i)$ and $(X_j, T_j)$ are identically distributed. The concordance index has been widely used in the survival setting as it can be corrected to account for censoring by only concerning itself with the global *ranking* capabilities of the model more than the accuracy of the predictions. While incomplete and criticizable, the concordance metric represents one important aspect of the model for practitioners: how accurate are the relative risks (see [30]). This is particularly important in the medical or financial setting where whether to select (for treatment or for financing) an individual over another is the useful actionable insight.

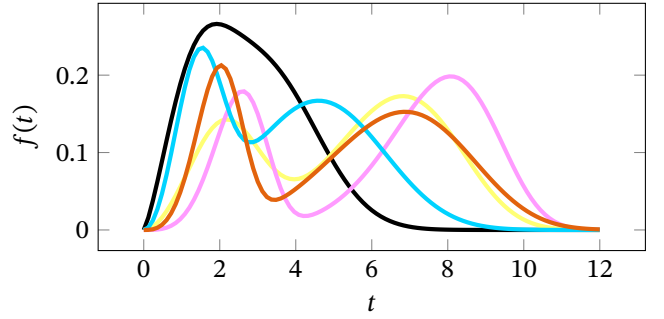

Figure 4. Different synthetic distributions for $d = 10$

While proportional hazard methods have a natural notion of risk score to compute the concordance index, this is not the case for our method or survival forests. In the random survival forest setting [31] the authors construct the risk score as

$$s(X_i) = \sum_{i=1}^{m} \hat{H}_i(t_i \mid X_i),$$

where $(t_i)_{i=1,...,m}$ are the unique event times in the training set. While a similar approach can be used for our model, we instead exploit the fact that we can easily and cheaply generate conditional observations to directly learn the ranking implied by the concordance. Given a test dataset $(Y_i, \delta_i, X_i)_{i=1,...,n}$ we first generate $n \times m$ observations $T_i = G_\theta(Z_k, X_i)$ with $Z_k \sim \mathcal{Z}$ i.i.d and define the score vector $\mathbf{s} = [s(X_1), ..., s(X_n)]$ as

$$s(X_i) = \frac{1}{m} \sum_j \mathbb{1}_{T_i > T_j}.$$

The competing approaches have been trained using the `PySurvival` [32] library and their concordance computed using the function scoring function from the same package. The results are presented in Table I.

Table I
CONCORDANCE ACHIEVED ON SYNTHETIC DATASETS.

| Method | Concordance |
| --- | --- |
| This work | **0.795691** |
| DeepSurv [2] | 0.762831 |
| Survival Forest [31] | 0.705942 |
| Cox PH | 0.666684 |

*B. Real Data*

We evaluate our approach compared to the state of the art on several open healthcare-related datasets as well as a proprietary internal banking dataset.

The four medical datasets considered are the Worcester Heart Attack Study (WHAS) [33], the Study to Understand Prognosis's Preferences Outcomes and Risks of Treatment (SUPPORT) [34], The Molecular Taxonomy of Breast Cancer International Consortium (METABRIC) [35] as well as the Rotterdam & German Breast Cancer Study Group (RGBSG) [36], [37]. The characteristics of the different datasets are summarized in Table II

Table II
DESCRIPTIVE STATISTICS OF THE REAL DATASETS USED IN THIS WORK.

| Dataset | $\mathbb{E}[\delta]$ | $d$ | $n_{\text{train}}$ | $n_{\text{test}}$ |
| --- | --- | --- | --- | --- |
| Metabric | 0.579307 | 9 | 1523 | 381 |
| RGBSG | 0.567652 | 7 | 1546 | 686 |
| Support | 0.680266 | 14 | 7098 | 1775 |
| WHAS | 0.421245 | 6 | 1310 | 328 |

The networks parameterizing $\sigma_\theta$, $\pi_\theta$, $f_\theta$ are chosen as simple feed forward neural networks, but as $\pi_\theta$ is time independent, a more expressive network can be chosen such as a transformer network [38] with a dense last layer if the covariates include unstructured text. Solving the neural differential equation involves repeated evaluation of the functional defining the dynamics; while solvers such as Tsit5 [39] and BS5 [40] are adaptive and only require a limited number of evaluations depending on the stiffness of the problem, we still keep the computational complexity of the time components low in order to keep training times low. As the accuracy of the solution is not of the utmost importance in our application (we are only interested in the generalization error, not the approximation error), we found it possible to use low accuracy solvers with high tolerances without loss of predictive performance.

The parameters used for the survival normalizing flows are summarized in Table V-B (with $L$ the number of layers and $S$ their size), while the parameters and results from the other techniques are taken as-is from their respective papers. The activation function used is SELU [41] for all datasets and we use the identity function as a last layer for the covariate networks $\pi$ and the softmax function for the $\sigma$ and $\theta$.

The performance of the different methods on the four datasets is summarized in Table IV. We see that Normalizing Flows outperform the state-of-the-art on 2 of the 4 datasets and only underperform compared to random forests on the WHAS dataset. Such a result is not unexpected: the covariates include highly engineered binary variables that were highly suspected to be indicators of future heart problems by the instigators, it is therefore expected (provided that their hypothesis was correct) that a space partitioning algorithm would perform close to optimally.

Table III
HYPERPARAMETERS SELECTED FOR THE SURVIVAL FLOWS USED IN TABLE IV.

| Dataset | $S_\pi$ | $S_\sigma$ | $S_g$ | $L_\pi$ | $L_\sigma$ | $L_g$ | K |
| --- | --- | --- | --- | --- | --- | --- | --- |
| Support | 4 | 4 | 12 | 3 | 3 | 3 | 16 |
| WHAS | 12 | 8 | 12 | 4 | 4 | 3 | 32 |
| RGBSG | 4 | 4 | 12 | 3 | 3 | 3 | 16 |
| Metabric | 4 | 4 | 12 | 3 | 4 | 4 | 32 |

Table IV
CONCORDANCE OF SURVIVAL FLOWS COMPARED TO COMPETING TECHNIQUES ACHIEVED ON MULTIPLE REAL DATASETS.

| | Concordance | | | |
| --- | --- | --- | --- | --- |
| Method | Support | WHAS | RGBSG | Metabric |
| This work | 0.61678 | 0.86059 | **0.68464** | **0.64879** |
| DeepSurv [2] | **0.61831** | 0.86262 | 0.66840 | 0.64337 |
| Survival Forest [31] | 0.61302 | **0.89362** | 0.65119 | 0.62433 |
| Cox PH | 0.58287 | 0.81762 | 0.65775 | 0.63062 |

*C. Portfolio Optimization by Simulation*

One significant advantage of our method is the ability to efficiently generate samples of $T$, the duration of interest; enabling the possibility to estimate higher order statistics that may depend on $T$ through a non-trivial process. We present here a toy example of such an application in order to motivate this characteristic.

We consider a synthetic dataset of financial entities representing a credit portfolio. For each client $i$ it is possible to buy an insurance (potentially on a fraction $\omega_i$) of duration $d_i$ (maximum time of the insurance), for a price $p_i$. If the client defaults during the contract duration ($T_i \leq d_i$) then the default is entirely covered, if not then a loss $l_i$ is incurred. If we define the portfolio loss as

$$L(\omega) = \sum_{i=1}^{K} \omega_i (\mathbb{1}_{t_i \leq d_i} - p_i),$$

We then want to minimize some metric of the risk incurred. Many such metrics exist but we chose here to optimize the *expected shortfall*, defined by

$$ES_\alpha(\omega) = \int_{-\infty}^{\alpha} F_{L(\omega)}^{-1}(\gamma) d\gamma.$$

This measure is equivalent in the continuous case to the tail conditional expectation and can be seen as *minimizing the expected extremal losses*. This objective is also desirable in many other fields such as predictive maintenance or industrial reliability, where minimizing the extreme defects is of particular interest. Not only the previous quantity can be estimated by Monte Carlo simulations using the normalizing flow learned by our method, but it is also possible to directly minimize the previous quantity by solving the convex optimization program

$$\arg\min_{\omega} \int_{-\infty}^{\alpha} F_{L(\omega)}^{-1}(\gamma) d\gamma.$$

which can be rewritten as the linear optimization program (see [42]), if we add constraints on the size of the portfolio $T$ as well as the size of the positions:

$$\arg\min_{\omega,\beta} \quad \beta + \frac{1}{1-\alpha} \int [L(\omega) - \beta]^+ p_L(y) dy \quad (8)$$
$$\text{s.t.} \quad 0 \leq \omega \leq 1$$
$$\omega^\intercal p = T$$

For simplicity we set here all durations to the same value $d_i = d$ and generate default events according to a simplified version of the law introduced here in subsection V-A: we sample 10, $(\tilde{X}_k) \sim \mathcal{U}([0,1]^{10})$, the 10 resulting vectors are then perturbated to form 200 feature vectors: $X_i = \tilde{X}_k + 0.1\varepsilon_i$ with $\varepsilon \sim \mathcal{N}(0, \mathbb{I}_{10})$. The time to defaults and censoring variables themselves still follow the model given previously. This simplified model represents the usually assumed *classes* of risk, such as industries, countries or intrinsic rating, which are usually assumed to be similar in terms of default.

Prices are chosen as 'fair prices' i.e. as the prices necessary to mitigate the expected losses $p_i = \mathbb{E}[\mathbb{1}_{t_i \leq d_i}]$. For the pricing we use separate Weibull models learned independently for each $k \in [1, \cdots, 10]$, a common practice in credit rating. We minimize (8) using the reference Weibull models and our model as estimators of the distribution of defaults. The real value of the minimum obtained by both methods is then obtained using the true (unknown) distribution to compute the true realized loss of the portfolio.

The true expected shortfall of the standard optimal portfolio is 8.1452 while the optimal portfolio formed using the survival flow estimates achieves an expected shortfall of −0.314. Purely on this metric, the more granular and accurate samples from the normalizing flow model results in a significantly better objective value. While the expected shortfall is a good objective with interesting mathematical characteristics as well as real economic interpretation, investors are in the end interested in the real possible returns. As seen in Figure 5, the potential losses from the optimal portfolio obtained by means of minimizing the expected shortfall derived the reference distribution, i.e. the distribution used for the pricing, are significantly higher than those obtained using samples from a normalizing flow based estimate of the survival distribution. In this toy example, we clearly observe better losses (or gains depending on the point of view) when sampling from the survival flow distribution. We expect this pattern to hold true for any complex decision process relying on accurately drawing random samples, such as experiments from queuing theory or integration from complex quantities such as in ray tracing.

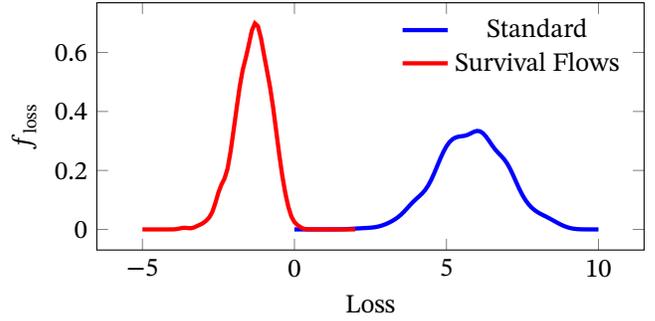

Figure 5. Realized losses for the optimal portfolios using the reference and normalizing flows estimates.

## VI. Conlusion

We introduced a novel conditional survival density estimator based on continuous normalizing flows and showed that it is able to outperform state-of-the art technique for a moderate increase in computational complexity. We show using a toy example designed to resemble a real-world problem commonly faced in finance that the ability to efficiently sample from the learned distribution can be incredibly valuable and more than offset the computational increase incurred during the learning phase. While the training runtime of our model can probably be significantly improved by carefully optimizing the code up to the standard of the already existing competing mature libraries, we believe that more research is necessary in order to achieve the best possible performance. It is known in the regression setting that augmenting the ODE [43] leads to significantly improved predictive and computational performance but those results cannot be directly applied to the normalizing flow setting. Similarly, methods that try to control the stiffness of the ODE such as [44] have shown great promises.